\documentclass[twoside,11pt]{article}
\usepackage{jmlr2e}

\usepackage[utf8]{inputenc}
\usepackage[T1]{fontenc}
\usepackage{url}
\usepackage{booktabs}
\usepackage{amsfonts}
\usepackage{nicefrac}
\usepackage{microtype}
\usepackage{xcolor}
\usepackage{xspace}

\usepackage{amsmath}
\usepackage{amssymb}
\usepackage{mathtools}
\usepackage{bm}

\usepackage{algorithm}
\usepackage{algpseudocode}

\usepackage{graphicx}
\usepackage{subcaption}
\usepackage{tikz}
\usetikzlibrary{arrows.meta, positioning, calc, shapes.geometric, fit, decorations.pathreplacing}

\usepackage{multirow}
\usepackage{makecell}

\graphicspath{{figures/}}

\newcommand{\R}{\mathbb{R}}
\newcommand{\E}{\mathbb{E}}
\newcommand{\Var}{\mathrm{Var}}
\newcommand{\Tr}{\mathrm{tr}}
\newcommand{\N}{\mathcal{N}}
\newcommand{\Lcal}{\mathcal{L}}
\newcommand{\Dcal}{\mathcal{D}}

\newcommand{\Icoarsefine}[2]{\mathcal{I}_{#1 \to #2}}

\newcommand{\uLF}{u_{\mathrm{LF}}}
\newcommand{\uHF}{u_{\mathrm{HF}}}
\newcommand{\SLF}{S_{\mathrm{LF}}}
\newcommand{\SHF}{S_{\mathrm{HF}}}

\newcommand{\vtheta}{v_\theta}

\newcommand{\Sgmd}{\Sigma_\delta}

\newcommand{\norm}[1]{\left\|#1\right\|}

\usepackage{cleveref}

\begin{document}
\firstpageno{1}

\title{Multi-Fidelity Flow Matching:\\Cascaded Refinement of PDE Solutions}

\author{%
\name Sipeng Chen \email sc25bg@fsu.edu \\
\addr Department of Computer Science, Florida State University
\AND
\name Junliang Liu \email jliu@eng.famu.fsu.edu \\
\addr Department of Materials Science and Engineering, Florida State University
\AND
\name Hewei Tang \email hewei.tang@austin.utexas.edu \\
\addr Hildebrand Department of Petroleum and Geosystems Engineering, The University of Texas at Austin
\AND
\name Shibo Li \email shiboli@cs.fsu.edu \\
\addr Department of Computer Science, Florida State University
}

\begingroup
\renewcommand\thefootnote{\fnsymbol{footnote}}
\maketitle
\endgroup


\begin{abstract}
The source distribution in conditional flow matching is a design parameter that can be calibrated to data, not a default isotropic prior. We exploit this in Multi-Fidelity Flow Matching (MFFM), a cascade refinement framework for parametric PDE solutions: the source is calibrated to the empirical low-to-high-fidelity residual scale with local Gaussian-blur correlation, and the velocity network is conditioned on the low-fidelity solution. Conditioning makes the residual refinement problem substantially easier than unconditional field generation, while residual-calibrated source noise improves the flow-matching training geometry. A multi-resolution cascade applies the same construction independently between adjacent fidelities. After level-wise flow-matching pretraining, we fine-tune the composed cascade end-to-end with a deterministic one-step rollout, which makes one velocity evaluation per cascade level the optimized operating point at inference. The result is a learned analog of multigrid refinement that reaches the finest grid in $L$ deterministic network evaluations per query. We validate MFFM on eight benchmarks: two super-resolution problems and six spatiotemporal forecasting tasks from PDEBench, The Well, and the FNO Navier--Stokes dataset.
\end{abstract}

\section{Introduction}
\label{sec:intro}

Solving a parametric PDE from its coefficients, initial conditions, or boundary conditions remains costly even with modern neural surrogates. The operator $\xi \mapsto \uHF$ is high-dimensional, geometry-sensitive, and unstable on parameter regimes far from training. Refining an already-computed approximate solution is a different and strictly easier task: the residual carries far less variance than the field itself. A coarse-mesh solver, a reduced-order model, or a neural-operator surrogate already captures the macroscopic structure of the high-fidelity field; the residual $\delta = \uHF - \uLF$ then has smaller variance and is concentrated in the directions where the cheap solver loses resolution. For typical parametric PDEs the LF and HF solutions are strongly correlated across most of the domain. Classical multi-fidelity methods \citep{kennedy2000predicting, perdikaris2017nonlinear, peherstorfer2018survey} and recent neural multi-fidelity surrogates \citep{lu2022multifidelity, howard2023multifidelity, li2022infinite, li2024multiresolution} have long exploited this regime.

\begin{figure}[t]
\centering
\includegraphics[width=0.77\linewidth]{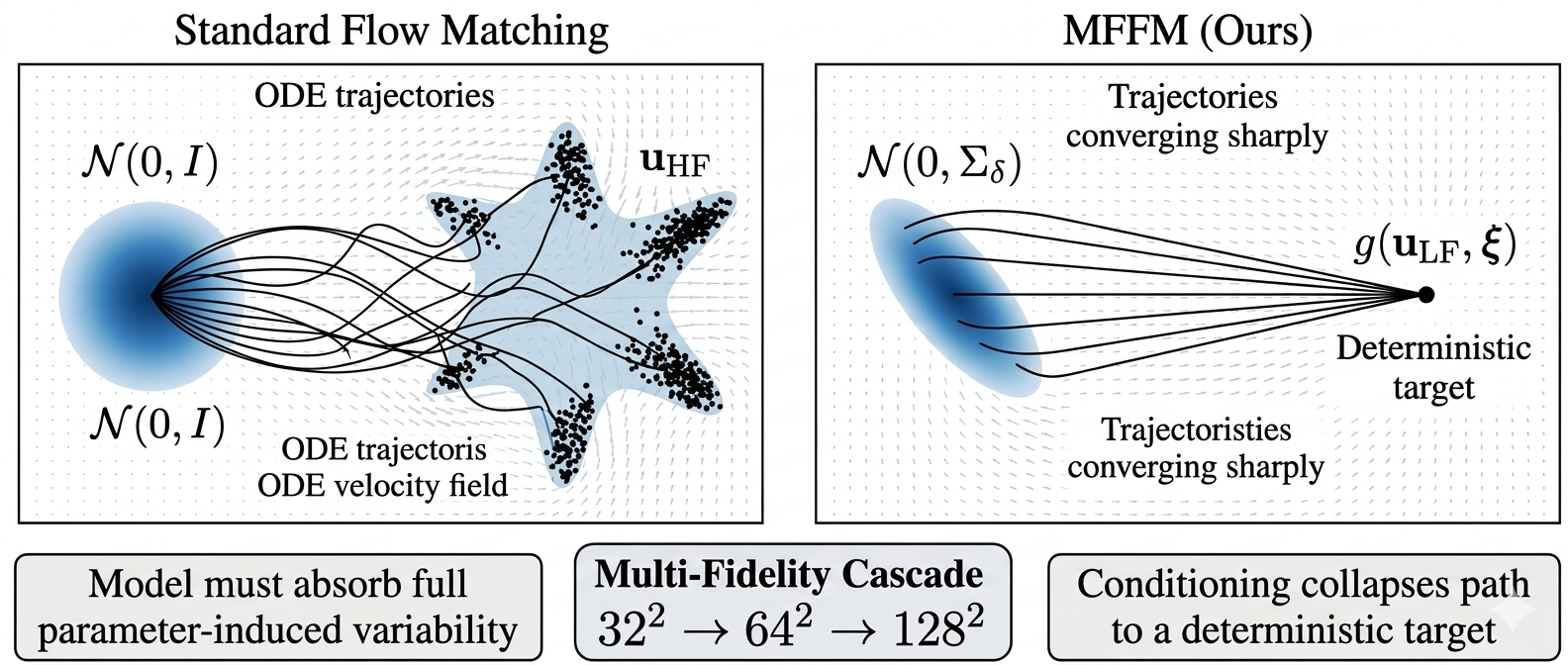}
\caption{Standard flow matching (\textit{left}) transports an isotropic source $\N(0, I)$ to the parameter-induced distribution $\uHF$ along curving ODE trajectories, so the model must absorb the full parametric variability of the target. MFFM (\textit{right}) replaces the source with one calibrated to the LF$\to$HF residual and conditions on the low-fidelity solution, making the residual refinement substantially easier than unconditional field generation. The full method composes $L$ such per-level refinements across nested resolutions as a cascade (\textit{bottom}; Sec.~\ref{sec:method-cascade}); after end-to-end fine-tuning (Sec.~\ref{sec:method-e2e}) one velocity evaluation per cascade level is the optimized operating point at inference.}
\label{fig:teaser}
\end{figure}

Multi-Fidelity Flow Matching (MFFM) sits downstream of operator-learning surrogates rather than competing with them. A neural operator absorbs the parametric variability of the solution manifold; MFFM consumes its output, or any low-fidelity solver's output, on a finer grid and produces a high-fidelity refinement through a flow-matching cascade. In practice we expect MFFM to follow a learned LF surrogate: the operator handles the parameter-to-solution map at its native resolution, and the cascade refines to the high-fidelity grid. Figure~\ref{fig:teaser} previews the geometric distinction between standard flow matching and the adapted-source variant.

We pose the LF-to-HF refinement as conditional residual flow matching. Generative approaches to PDE-related problems \citep{huang2024diffusionpde, lippe2024pde, bastek2025physics, shu2023physics, baldan2025pbfm, hou2025cfo} typically take the source distribution as uninformative noise, so the model must recover the macroscopic component of $\uHF$ at every ODE step and pays the same transport cost as single-fidelity generation. We instead calibrate the source to empirical residual statistics: in the main implementation, iid Gaussian noise is locally correlated by Gaussian blur, normalized samplewise, and scaled by the empirical per-coordinate residual standard deviation. The velocity network is conditioned on $\uLF$ at every layer. The two design choices compose: (i) the residual-calibrated source reduces the marginal scale gap between source and target and better matches the local structure of PDE residuals; (ii) conditioning on $\uLF$ makes the residual refinement target substantially more concentrated than the full high-fidelity field. After level-wise flow-matching pretraining, we fine-tune the composed cascade end-to-end using a deterministic one-step rollout. Thus one midpoint velocity evaluation per cascade level is the optimized operating point used for all main predictions.

Stacking the same construction across nested resolutions $G_0 \subset G_1 \subset \cdots \subset G_L$ produces a learned analog of multigrid refinement that reaches the finest grid in $L$ deterministic network forward passes per query, with per-level networks correspondingly smaller because each handles a narrower band of residual frequencies.

\section{Background}
\label{sec:background}

The classical multi-fidelity treatment of \citet{kennedy2000predicting} models the high-fidelity output as a scaled low-fidelity output plus a Gaussian-process discrepancy, and motivates the additive decomposition we adopt. Nonlinear couplings \citep{perdikaris2017nonlinear} replace the scaling with a learned map; \citet{peherstorfer2018survey} surveys two decades of multi-fidelity methodology across uncertainty propagation, optimization, and inverse problems. An earlier surrogate-modelling thread \citep{forrester2007multifidelity} formalized cheap-to-expensive corrections in design optimization. Neural realizations of multi-fidelity correction include multi-fidelity DeepONets \citep{lu2022multifidelity, howard2023multifidelity}, infinite-fidelity coregionalization \citep{li2022infinite}, and active-learning-driven multi-resolution operator architectures \citep{li2024multiresolution, li2022deep, li2020multifidelity}, all built on the broader neural-operator framework surveyed by \citet{kovachki2023neural}. These approaches model the LF-to-HF correction as a deterministic regressor; MFFM instead uses residual flow matching with a source distribution calibrated to empirical LF-to-HF residual statistics, followed by deterministic cascade fine-tuning for point prediction.

Conditional flow matching \citep{lipman2023flow}, a continuous-time generative-modelling framework descended from neural ODEs \citep{chen2018neural} and the diffusion family \citep{ho2020denoising, song2021scorebased}, trains a velocity network by regressing onto closed-form per-sample conditional velocities, avoiding ODE simulation at training. Under the linear interpolation path $x_t = t\,x_1 + (1-t)\,x_0$, the per-sample conditional velocity reduces to the constant $x_1 - x_0$. Rectified flow \citep{liu2023flow} derives the same path from a trajectory-straightening argument and can support low-NFE inference when the source-target coupling is well learned. Minibatch optimal-transport coupling \citep{tong2024improving} further reduces regression variance. Across these variants the source is usually treated as a fixed reference distribution. Bridge methods \citep{debortoli2021bridge, shi2024bridge} and stochastic interpolants \citep{albergo2023stochastic} establish the broader formal framework within which non-noise sources are admissible.

Cascaded diffusion \citep{ho2022cascaded, saharia2022photorealistic} composes denoising samplers across resolutions, and multi-level Monte Carlo \citep{giles2015multilevel} achieves an analogous variance reduction in the deterministic regime; both inform the cascade design here. Image-domain cascades typically retain a multi-step sampler at every level, whereas MFFM uses adjacent-fidelity residual flow matching as pretraining and then operates as a one-forward-pass-per-level deterministic refiner after end-to-end cascade fine-tuning.

\section{Multi-fidelity flow matching}
\label{sec:method}

\subsection{Problem formulation}
\label{sec:method-setup}

Let $\Xi \subset \R^p$ index a family of parametric PDEs $\Lcal_\xi[u] = f_\xi$ on a fixed domain $\Omega \subset \R^d$. A low-fidelity solver $\SLF$ produces $\uLF = \SLF(\xi)$ at cost $c_L$ on a coarse grid, and a high-fidelity solver $\SHF$ produces $\uHF = \SHF(\xi)$ at cost $c_H \gg c_L$ on a fine grid.

Our primary setting is \emph{spatiotemporal PDE forecasting}: $\uLF$ stacks $T$ past frames at the coarse grid into channels, $\uHF$ stacks the corresponding $T$ future frames at the fine grid, and refinement runs on the HF grid after upsampling $\uLF$. This is the PDEBench setup of Section~\ref{sec:experiments}. The construction also covers \emph{spatial or space--time super-resolution} ($T = 1$ for spatial fields), and we include Darcy and Burgers as such validations.

Training assumes a paired dataset $\Dcal = \{(\uLF^{(i)}, \uHF^{(i)})\}_{i=1}^N$ obtained by running both solvers or constructing paired LF--HF fields on identical PDE instances. The parameter $\xi$ generates the data but is not provided to the model.

The setting differs from operator learning in two consequential ways. Operator surrogates approximate $\xi \mapsto \uHF$ from $\xi$ alone and must absorb every source of parametric variability without a solution-side prior. MFFM operates downstream of $\SLF$ and exploits $\uLF$ as an informative summary of $\uHF$; the map we learn is the conditional refinement $\uLF \mapsto \uHF$, which removes the dominant macroscopic component of the parametric variation. The two settings compose: a learned operator (FNO, DeepONet, multi-fidelity DeepONet, COMPOL \citep{sun2025compol, li2024multiresolution}) can play the role of $\SLF$, and the resulting refinement provides high-fidelity predictions without re-running the expensive HF solver.

We pose the refinement as conditional residual flow matching, learning a velocity model on the residual $\delta = \uHF - \uLF$ conditioned on $\uLF$.

\subsection{Adapted-source flow matching}
\label{sec:method-residual}

The construction is best understood by contrast. Standard flow matching transports samples from an uninformative source distribution to the data and pays a transport cost proportional to the geometric gap between the two. In the multi-fidelity setting we instead draw the source from empirical residual statistics, so the source and target share the same residual scale from the outset; the network learns the conditional coupling between source and target rather than rescaling between distributions of incompatible scale.

We work directly on the residual $\delta = \uHF - \uLF \in \R^n$ and define a flow-matching generator with a calibrated residual source. In the main implementation, the source is a locally correlated Gaussian-blur residual source:
\begin{equation}
z \sim \N(0,I), \qquad
\tilde z =
\frac{K_\tau * z}{\widehat{\mathrm{Std}}(K_\tau * z) + \epsilon_{\mathrm{num}}}, \qquad
\varepsilon = \sigma_\delta \odot \tilde z,
\qquad
\sigma_j^2 = \widehat{\Var}_{\Dcal}[\delta_j],
\label{eq:source}
\end{equation}
where $K_\tau$ is a depthwise Gaussian blur kernel, $*$ denotes spatial convolution, and the standardization is applied samplewise. This construction preserves the empirical per-coordinate residual scale while introducing local spatial correlation, matching the structure of PDE residual fields more closely than pixelwise independent noise. The calibrated diagonal source $\varepsilon \sim \N(0,\mathrm{diag}(\sigma_\delta^2))$ is included as an ablation.

Following \citet{lipman2023flow}, we adopt the linear interpolation conditional path $\delta_t = t\delta + (1-t)\varepsilon$, $t \in [0,1]$, which has constant per-sample conditional velocity $v(\delta_t, t \mid \delta, \varepsilon) = \delta - \varepsilon$. The MFFM regression objective
\begin{equation}
\mathcal{J}_{\mathrm{MFFM}}(\theta) = 
\E_{t \sim \mathcal{U}[0,1]}\,
\E_{(\uLF, \delta) \sim \Dcal}\,
\E_{\varepsilon \sim q_\delta}
\norm{\vtheta(\delta_t, t, \uLF) - (\delta - \varepsilon)}^2
\label{eq:loss}
\end{equation}
exposes $\uLF$ to the velocity network at every layer rather than through a single conditioning bottleneck, so $\vtheta$ can learn the conditional refinement structure rather than the unconditional residual marginal. Here $q_\delta$ denotes the calibrated residual source in Eq.~\eqref{eq:source}. At inference time, one can sample $\varepsilon \sim q_\delta$ and integrate the inference ODE
\begin{equation}
\frac{d\delta_t}{dt} = \vtheta(\delta_t, t, \uLF^*), \qquad \delta_0 = \varepsilon,
\label{eq:ode}
\end{equation}
from $t = 0$ to $t = 1$, returning $\hat{\uHF} = \uLF^* + \delta_1$. Drawing $K$ independent source samples produces an optional stochastic ensemble. For the deterministic point-estimate results in Table~\ref{tab:main}, we use the end-to-end fine-tuned deterministic cascade described in Section~\ref{sec:method-cascade}; stochastic source sampling is not used for the main accuracy results.

The choice of source in Eq.~\eqref{eq:source} is a normalization and correlation-design decision (an analog of the noise-schedule design space catalogued for diffusion models by \citet{karras2022elucidating}), not the mechanism that makes one-step refinement effective. Matching the empirical residual scale aligns the marginal magnitude of $\delta_t$ with both endpoints throughout the path; adding Gaussian-blur correlation further matches the local smoothness of PDE residuals. Under the diagonal source, the transport scale satisfies $\E\norm{\delta - \varepsilon}^2 = 2\Tr(\Sgmd)$ with $\Sgmd=\mathrm{diag}(\sigma_\delta^2)$; on Darcy at $128^2$ this is roughly one-tenth of the corresponding $\Theta(n)$ value an isotropic $\N(0, I)$ source would induce on unit-normalized fields (Appendix~\ref{app:transport-variance}). The mechanism making one-step residual refinement effective is conditioning on $\uLF$, formalized in Section~\ref{sec:method-onestep}.

\subsection{Single-step inference under deterministic conditioning}
\label{sec:method-onestep}

By default the inference ODE \eqref{eq:ode} may require many integration steps. The training loss in Eq.~\eqref{eq:loss} drives $\vtheta$ toward the marginal Bayes-optimal velocity $v^*(z, t) = \E[\delta - \varepsilon \mid \delta_t = z, t]$, which is generally $t$-dependent: for two independent zero-mean Gaussians with equal covariance, $v^*(z, t) = (2t-1)/(2t^2 - 2t + 1)\,z$. Universal single-step claims for conditional flow matching are therefore false. What MFFM uses is the conditional regressor under $\uLF$, which becomes simple for a structural reason specific to the multi-fidelity setting.

\paragraph{One-step inference under deterministic conditioning.}
Condition on $\uLF$ and assume that the LF solver is injective on the parameter space, so that $\uLF$ uniquely determines $\xi$ and, since $\SHF$ is deterministic, also determines $\uHF$. The conditional law of $\delta$ given $\uLF$ is then the Dirac measure concentrated at the deterministic refinement target $g(\uLF) := \uHF - \uLF$. From this:
\begin{equation}
\delta_t = t\,g(\uLF) + (1-t)\,\varepsilon
\end{equation}
is a deterministic function of $\varepsilon$ given the conditioning, and the source is recoverable from the input via $\varepsilon = (\delta_t - t\,g)/(1-t)$ for $t < 1$. Substituting into the regression target yields the conditional Bayes-optimal regressor
\begin{equation}
v^*(z, t \mid \uLF) = (g(\uLF) - z)/(1 - t),
\label{eq:cond-vel-formula}
\end{equation}
which is generically $t$-dependent off the inference path but reduces to the constant $g - \varepsilon$ along it. A single Euler step from $\delta_0 = \varepsilon$ would then return $\delta_1 = \varepsilon + (g - \varepsilon) = g(\uLF)$, the high-fidelity refinement target, \emph{if} the trained network coincided with the Bayes regressor.

In practice $\hat{g}_\theta$ approximates $g$ to a residual error we measure empirically. After deterministic end-to-end fine-tuning, the one-step rollout becomes the optimized operating point rather than a coarse approximation to a many-step ODE solve; Section~\ref{sec:exp-nfe} shows that increasing NFE does not improve the fine-tuned predictor. We do not claim exactness; we claim the construction makes one step the appropriate operating point under (i) $\SHF$ deterministic and $\SLF$ sufficiently informative on $\Xi$, the standard PDE-solver setting; (ii) source independence $\varepsilon \perp \uLF$, which holds by construction; and (iii) evaluation away from $t=1$, avoiding the $1/(1-t)$ singularity in Eq.~\eqref{eq:cond-vel-formula}. The deterministic e2e predictor uses the midpoint $t=1/2$, as described in Section~\ref{sec:method-cascade}.

\paragraph{Empirical confirmation.}
Section~\ref{sec:exp-nfe} evaluates the fine-tuned deterministic cascade with multiple Euler evaluations per level. Increasing NFE does not improve NRMSE, confirming that the one-step rollout is the appropriate operating point for the e2e fine-tuned predictor.

\subsection{Multi-level cascade}
\label{sec:method-cascade}

Two regimes break the single-shot argument of Section~\ref{sec:method-onestep}. Stochastic $\SHF$ (random forcing, sub-grid Monte Carlo, SPDEs) and non-injective $\SLF$ replace the Dirac measure with a non-degenerate conditional distribution; this regime does not arise for the deterministic numerical solvers we consider. Insufficient network capacity, by contrast, is the operative bottleneck whenever the LF-HF gap is large: a single-shot refinement asks one network to recover $g$ across the full gap, where $g$ has large per-coordinate variance and the conditional structure is harder to learn at fixed parameter budget. The cascade addresses this regime by splitting the bottom-to-top refinement into $L$ adjacent-fidelity steps; each per-level $g_\ell$ has smaller variance and stronger conditioning than the bottom-to-top $g$, and the conditional argument above applies more cleanly per level than per single shot.

Let nested grids $G_0 \subset G_1 \subset \cdots \subset G_L$ form a multi-resolution hierarchy with prolongation operators $\Icoarsefine{\ell}{\ell+1}: \R^{n_\ell} \to \R^{n_{\ell+1}}$ (bilinear interpolation in our experiments). At each level the cascade applies one MFFM step:
\begin{equation}
u_{\ell+1} = \Icoarsefine{\ell}{\ell+1}(u_\ell) + \delta_\ell, \qquad \ell = 0, \ldots, L-1,
\label{eq:cascade-decomp}
\end{equation}
where $\delta_\ell$ is generated by an MFFM model trained on level-$\ell$ pairs $(\Icoarsefine{\ell}{\ell+1}(u_\ell),\, u_{\ell+1})$. Eq.~\eqref{eq:cascade-decomp} defines the cascade additive decomposition. Each level has its own velocity network $v_{\theta_\ell}$ trained with the per-level objective
\begin{equation}
\mathcal{J}_\ell(\theta_\ell) = \E_{t,\, \varepsilon_\ell,\, (u_\ell, \delta_\ell)} \norm{v_{\theta_\ell}(\delta_{\ell, t}, t, \tilde{u}_\ell) - (\delta_\ell - \varepsilon_\ell)}^2,
\label{eq:cascade-loss}
\end{equation}
with $\tilde{u}_\ell = \Icoarsefine{\ell}{\ell+1}(u_\ell)$, $\varepsilon_\ell \sim q_{\delta_\ell}$, and $\delta_{\ell,t}=t\delta_\ell+(1-t)\varepsilon_\ell$, where $q_{\delta_\ell}$ is the per-level calibrated residual source from Eq.~\eqref{eq:source}.

\paragraph{Per-level deterministic conditioning.}
Training each level with the calibrated-source loss in Eq.~\eqref{eq:cascade-loss}, the conditioning argument of Section~\ref{sec:method-onestep} transfers level by level: each network observes the prolongated state $\tilde u_\ell$ and only needs to predict the adjacent-fidelity residual $u_{\ell+1}-\tilde u_\ell$. During level-wise pretraining, the conditioning states come from ground-truth intermediate fidelities; during inference, they are produced by previous cascade levels. This creates a cascade covariate shift. We address it directly with the deterministic end-to-end fine-tuning stage below, which optimizes the composed one-step rollout used at test time.

\paragraph{Per-level transport-cost bound.}
The per-level residual variance can be related to the empirical adjacent-fidelity correlation $\rho_{\ell, \ell+1, j} = \mathrm{Corr}(u_{\ell, j}, u_{\ell+1, j})$. Treating adjacent fidelities as having approximately equal per-coordinate variance $\sigma_{\ell, j}^2 \approx \sigma_{\ell+1, j}^2$, the per-coordinate residual variance satisfies
\begin{equation}
\sigma_{\delta_\ell, j}^2 \approx 2(1 - \rho_{\ell, \ell+1, j})\,\sigma_{\ell+1, j}^2,
\label{eq:per-level-var}
\end{equation}
mirroring the bottom-to-top calculation in Appendix~\ref{app:transport-variance}. Adjacent fidelities are typically more correlated than non-adjacent ones in our benchmarks, so the adjacent-level residual scale is smaller than the direct bottom-to-top residual scale, making each per-level transport task easier than the single-shot coarsest-to-finest task.

\paragraph{Cost and parameter allocation.}
Total cascade inference cost is $L$ forward passes per query. Higher-resolution levels handle progressively smaller corrections (Eq.~\eqref{eq:per-level-var} with $\rho_{\ell, \ell+1}$ closer to one for finer level pairs), so the velocity network at finer levels uses fewer parameters per coordinate. The variance-allocation parallel with multi-level Monte Carlo \citep{giles2015multilevel} is direct: MLMC distributes samples adaptively across levels to minimize variance per dollar of compute, and the cascade allocates parameters across levels by the same logic.

\subsection{End-to-end cascade fine-tuning}
\label{sec:method-e2e}

The independent level-wise training objective in Eq.~\eqref{eq:cascade-loss} identifies each refinement model on the training chain induced by ground-truth intermediate fidelities. At inference time, however, the conditioning state at level $\ell$ is produced by the previous learned refinement models, so the per-level networks are evaluated outside the support seen during their independent training. We address this cascade-induced covariate shift directly with a second training stage: after independent level-wise pretraining, we unroll the full cascade from the coarsest input $u_0$ to the finest grid using one Euler evaluation per level and initialize the residual state at each level by zero:
\[
    \hat u_{\ell+1}
    =
    I_{\ell\to \ell+1}(\hat u_\ell)
    +
    v_{\theta_\ell}\!\left(
        0,\; \tfrac{1}{2},\; I_{\ell\to \ell+1}(\hat u_\ell)
    \right).
\]
All level networks $\{\theta_\ell\}_{\ell=0}^{L-1}$ are then optimized jointly against the final high-fidelity target with a relative $L_2$ loss,
\[
    \mathcal{L}_{\mathrm{e2e}}
    =
    \frac{\|\hat u_L - u_L\|_2}{\|u_L\|_2 + \epsilon}.
\]
This stage does not change the cascade architecture or the one-step inference budget; it adapts the independently trained per-level refiners to the distribution of conditioning states they actually encounter when composed, and turns the one-step rollout into the optimized operating point rather than a coarse approximation to a many-step ODE solve. Section~\ref{sec:exp-nfe} confirms empirically that increasing the per-level NFE on the fine-tuned cascade does not improve NRMSE. The calibrated source additionally defines an optional stochastic sampling interface (sample $\varepsilon_\ell \sim q_{\delta_\ell}$ at each cascade level and integrate the learned velocity field) that we use only as a diagnostic of predictive spread; a stochastic end-to-end variant is reported in Table~\ref{tab:stochastic-e2e} (Appendix~\ref{app:stochastic-e2e}).

\subsection{Connection to multigrid and inference cost}
\label{sec:method-multigrid}

The cascade has a direct analogy to multigrid in classical numerical analysis. In multigrid, a coarse-to-fine V-cycle alternates relaxation steps at multiple grid resolutions to accelerate iterative-solver convergence: the coarsest grid handles low-frequency error modes that fine-grid relaxation cannot reach efficiently, and successive finer grids correct progressively higher-frequency components. MFFM's cascade is a learned refinement analog: each level corrects residual frequencies that the prolongation from the coarser level fails to capture, and the deterministic e2e operating point replaces the relaxation step at each level with a single network forward pass. MFFM inherits multigrid's parameter efficiency: per-level networks can be smaller because each handles a narrower band of residual frequencies, and per-level training is the analog of independent local relaxation operators.

The total inference cost is $L \cdot c_{\mathrm{net}}$ per query, where $c_{\mathrm{net}}$ is one velocity-network forward pass and $L$ is the cascade depth. Compared with running the HF solver, the speedup is $c_{\mathrm{HF}} / (L \cdot c_{\mathrm{net}})$. This is favorable when the HF solver is significantly more expensive than the network, which holds for most parametric PDE workflows of practical interest: HF cost grows with grid resolution and physical complexity, often by orders of magnitude, while network cost is fixed by the architecture and grows slowly with input resolution.

\paragraph{Architecture (details in Appendix~\ref{app:impl}).}\label{sec:method-arch}
The velocity network is a U-Net-style residual convolutional network that takes the channel concatenation $(\delta_t, \uLF)$ as input and outputs a velocity field on the same lattice; the cascade prolongation $\Icoarsefine{\ell}{\ell+1}$ is bilinear. Architecture is not a contribution of this work and operator backbones (FNO, DeepONet, neural-operator transformers) can be substituted.

\section{Related work}
\label{sec:related}

\textbf{Multi-fidelity surrogates.} The classical multi-fidelity decomposition that MFFM adopts traces to \citet{kennedy2000predicting} and its nonlinear extensions \citep{perdikaris2017nonlinear}, surveyed in \citet{peherstorfer2018survey}. Classical Gaussian-process variants retain a posterior by construction but scale poorly to field-valued solutions; neural variants such as multi-fidelity DeepONets \citep{lu2022multifidelity, howard2023multifidelity}, infinite-fidelity coregionalization \citep{li2022infinite}, and multi-resolution operator architectures \citep{li2024multiresolution, li2022deep, li2020multifidelity} model the LF-to-HF correction as a deterministic regressor. Message-passing PDE solvers \citep{brandstetter2022message} pursue the deterministic-operator route on graph-structured discretizations.
\textbf{Generative refinement for PDE solutions.} Generative approaches to PDE problems \citep{huang2024diffusionpde, bastek2025physics, baldan2025pbfm, hou2025cfo} typically draw the source from uninformative noise. Closer super-resolution and forecasting precedents include physics-informed score-matching for inverse problems \citep{holzschuh2023inverse}, dynamics-informed diffusion forecasting \citep{cachay2023dyffusion}, and high-fidelity flow reconstruction \citep{shu2023physics}, each of which uses a noise source above a low-resolution conditioning observation. The closest conceptual neighbor PDE-Refiner \citep{lippe2024pde} inserts a diffusion-based refinement on top of an autoregressive prediction with the coarse predictor as a fixed denoising start, but does not calibrate the source to empirical LF-to-HF residual statistics.
\textbf{Cascade structure and source design.} The cascade structure draws on cascaded diffusion \citep{ho2022cascaded, saharia2022photorealistic} and the deterministic-regime analogs of multi-level Monte Carlo \citep{giles2015multilevel} and learned multigrid \citep{greenfeld2019learning}, but after deterministic end-to-end fine-tuning MFFM operates each level with a single forward pass. Among flow-matching variants that admit non-trivial source-target couplings, including multisample flow matching \citep{pooladian2023multisample}, bridge methods \citep{debortoli2021bridge, shi2024bridge}, and stochastic interpolants \citep{albergo2023stochastic}, none treats the source statistics (residual scale and local correlation) as data-estimable design parameters calibrated to a specific application; MFFM identifies multi-fidelity refinement as the regime where this calibration pays off, since the LF-HF data correlation directly determines a residual source design that makes deterministic cascade refinement effective.
\section{Experiments}
\label{sec:experiments}

\subsection{Setup}
\label{sec:exp-setup}

\textbf{Benchmarks.} We evaluate on eight benchmarks. Two are super-resolution problems: 2D Darcy flow with log-Gaussian permeability, and 1D Burgers' equation as a 2D space--time field with both axes in the resolution hierarchy. The remaining six are spatiotemporal forecasting tasks: Shallow Water (SW) and Diffusion Reaction (DR) from PDEBench~\citep{takamoto2022pdebench}; Shear Flow tracer (Shear-T), Shear Flow pressure (Shear-P), and Active Matter concentration (AM) from The Well~\citep{ohana2024well}; and 2D Navier--Stokes (NS) from the FNO dataset~\citep{li2021fourier}. For the forecasting tasks the input is a block of $T = 25$ past frames at the coarse grid and the target is the corresponding block of $T = 25$ future frames at the high-fidelity grid, with the temporal axis stacked as channels.
\textbf{Methods compared.} We compare nine methods. \textsc{Bilinear} performs no learning and upsamples $\uLF$ to the HF grid (Section~\ref{sec:exp-bilinear} discusses its role as a task-difficulty zero line). \textsc{FNO-direct} \citep{li2021fourier} and \textsc{DeepONet-direct} \citep{lu2021learning} take the PDE parameter $\xi$ as input and predict $\uHF$ directly. \textsc{FNO-residual} and \textsc{DeepONet-residual} take $\uLF$ as input and predict $\uHF$ with the same conditioning as MFFM but a deterministic operator backbone. \textsc{F-FNO} \citep{tran2023factorized} and \textsc{CNO} \citep{raonic2023convolutional} are recent neural operators in the same architectural family. \textsc{PDE-Refiner} \citep{lippe2024pde} is the closest probabilistic baseline (iterative diffusion-style refinement; $K = 4$ steps). \textsc{FM (single-level)} is the single-shot variant of MFFM with $L = 1$ at the largest parameter count we tested, isolating the cascade contribution. \textsc{MFFM (cascade)} is the method of Section~\ref{sec:method}.
\textbf{Metrics and protocol.} We report global NRMSE; all MFFM main numbers use the deterministic end-to-end cascade point predictor. All methods are trained with AdamW at learning rate $10^{-5}$, cosine decay, and gradient clipping at norm $1.0$. MFFM additionally uses EMA decay $0.999$ during per-level pretraining and an end-to-end fine-tuning stage at one fifth of the base learning rate, with checkpoint selection by validation NRMSE.

\subsection{Main results}
\label{sec:exp-main}

Table~\ref{tab:main} reports NRMSE on all eight benchmarks for the nine methods, split into the two super-resolution problems and the six spatiotemporal forecasting tasks.

\begin{table}[t]
\centering
\footnotesize
\caption{NRMSE on eight PDE benchmarks (mean $\pm$ std over $3$ seeds). For readability the table is split into two halves: top, Darcy, Burgers, SW, DR; bottom, Shear-T, Shear-P, AM, NS. Darcy and Burgers are spatial super-resolution problems; the other six are spatiotemporal forecasting ($T{=}25$ past LF frames $\to$ future HF frames). Bold marks the best method per benchmark. \textsc{FNO/DeepONet (direct)} take $\xi$ as input; all other learning methods take the LF solution.}
\label{tab:main}
\setlength{\tabcolsep}{4pt}

\resizebox{\textwidth}{!}{%
\begin{tabular}{lcccc}
\toprule
Method & Darcy & Burgers & SW & DR \\
\midrule
Bilinear              & 0.08616                       & 0.1541              & 0.1007                       & 3.170 \\
FNO (direct)          & 0.1994 $\pm$ 0.0041           & 0.1099 $\pm$ 0.022  & 0.05037 $\pm$ 0.0027         & 0.5557 $\pm$ 0.0047 \\
FNO (residual)        & 0.07051 $\pm$ 0.024           & 0.06595 $\pm$ 0.0057 & 0.03905 $\pm$ 0.0033        & 0.7910 $\pm$ 0.012 \\
DeepONet (direct)     & 0.2467 $\pm$ 0.011            & 0.6291 $\pm$ 0.0014 & 0.08304 $\pm$ 0.000042       & 0.9655 $\pm$ 0.0034 \\
DeepONet (residual)   & 0.06602 $\pm$ 0.00031         & 0.1269 $\pm$ 0.0010 & 0.1058 $\pm$ 0.0000040       & 1.8302 $\pm$ 0.0019 \\
F-FNO                 & \textbf{0.05830 $\pm$ 0.00040} & 0.08879 $\pm$ 0.016 & 0.03575 $\pm$ 0.00093        & 1.102 $\pm$ 0.18 \\
CNO                   & 0.8090 $\pm$ 0.42             & 0.1376 $\pm$ 0.028  & 0.1596 $\pm$ 0.020           & 3.215 $\pm$ 0.25 \\
PDE-Refiner           & 0.09255 $\pm$ 0.014           & 0.1439 $\pm$ 0.0040 & 0.1351 $\pm$ 0.00029         & 2.378 $\pm$ 0.020 \\
\midrule
\textbf{MFFM (cascade)} & 0.06198 $\pm$ 0.0011         & \textbf{0.02488 $\pm$ 0.00049} & \textbf{0.004756 $\pm$ 0.000076} & \textbf{0.2509 $\pm$ 0.0042} \\
\bottomrule
\end{tabular}}

\smallskip

\resizebox{\textwidth}{!}{%
\begin{tabular}{lcccc}
\toprule
Method & Shear-T & Shear-P & AM & NS \\
\midrule
Bilinear              & 0.6260              & 0.9011              & 0.003635                       & 0.6172 \\
FNO (direct)          & 0.3333 $\pm$ 0.0046 & 0.9372 $\pm$ 0.0045 & 0.9672 $\pm$ 0.012             & 0.1919 $\pm$ 0.0040 \\
FNO (residual)        & 0.3278 $\pm$ 0.0037 & 3.566 $\pm$ 0.18    & 0.02644 $\pm$ 0.0051           & 0.1952 $\pm$ 0.0041 \\
DeepONet (direct)     & 0.7080 $\pm$ 0.00021 & 0.9799 $\pm$ 0.011  & 0.007937 $\pm$ 0.00063        & 0.6104 $\pm$ 0.013 \\
DeepONet (residual)   & 0.5066 $\pm$ 0.0018 & 8.491 $\pm$ 0.0035  & 0.003441 $\pm$ 0.000012        & 0.4185 $\pm$ 0.0094 \\
F-FNO                 & 0.3161 $\pm$ 0.0074 & 2.550 $\pm$ 0.44    & 0.01311 $\pm$ 0.010            & 0.1984 $\pm$ 0.0067 \\
CNO                   & 0.3928 $\pm$ 0.0023 & 13.89 $\pm$ 1.7     & 0.09396 $\pm$ 0.015            & 0.2404 $\pm$ 0.010 \\
PDE-Refiner           & 0.5854 $\pm$ 0.0016 & 59.50 $\pm$ 3.2     & 0.09890 $\pm$ 0.00030          & 0.6125 $\pm$ 0.000055 \\
\midrule
\textbf{MFFM (cascade)} & \textbf{0.2253 $\pm$ 0.0012} & \textbf{0.2671 $\pm$ 0.0033} & \textbf{0.003162 $\pm$ 0.000017} & \textbf{0.04779 $\pm$ 0.00031} \\
\bottomrule
\end{tabular}}

\end{table}

MFFM-cascade achieves the best NRMSE on seven of the eight benchmarks. The exception is Darcy, where F-FNO obtains the lowest error ($0.05830$ vs.\ MFFM-cascade $0.06198$); this is also the benchmark where bilinear is already strong ($0.08616$), leaving less headroom for refinement.

Among the deterministic operator baselines, FNO and F-FNO are the strongest competitors but inconsistent across datasets: FNO-direct is competitive on Diffusion Reaction but fails on Active Matter, FNO-residual degrades sharply on Shear-P, and F-FNO wins Darcy but stays well behind MFFM-cascade on the harder tasks. The catastrophic-divergence pattern is most visible on Shear-P (bilinear NRMSE $> 0.9$): FNO-residual ($3.566$), CNO ($13.89$), and PDE-Refiner ($59.50$) all fail to fit, while MFFM-cascade obtains $0.2671$.

The largest absolute improvements occur where the LF-to-HF gap is large. On Shallow Water, MFFM-cascade reaches $0.004756$ vs.\ $0.03905$ for FNO-residual and $0.04761$ for single-level FM (Table~\ref{tab:ablation}); on Diffusion Reaction, $0.2509$ vs.\ $0.7910$ for FNO-residual; on Navier--Stokes, $0.04779$ vs.\ $\sim 0.20$ for the FNO variants and $0.3136$ for single-level FM (Figure~\ref{fig:ns-vis-app}, Appendix~\ref{app:ns-vis}, shows a qualitative example). The single-level ablation in Table~\ref{tab:ablation} isolates the cascade contribution: MFFM-cascade improves on single-level FM on every benchmark and by an order of magnitude on Burgers, SW, DR, and NS, supporting the claim that the adjacent-fidelity decomposition is structural rather than a side-effect of the residual flow-matching objective.

\subsection{Ablation Studies}
\label{sec:exp-ablation}

Table~\ref{tab:ablation} ablates the four main design choices of MFFM: the calibrated residual source (vs.\ IID noise of matched average variance), local Gaussian-blur correlation (vs.\ a diagonal source), the multi-resolution cascade (vs.\ a single-shot model), and residual flow matching (vs.\ flow matching on the high-fidelity field).

\begin{table}[t]
\centering
\small
\caption{Ablation of MFFM components (global NRMSE). ``w/o Gaussian blur'': diagonal source without local correlation. ``IID noise'': isotropic noise of matched average variance. ``MFFM-single'': one residual flow from coarsest to finest grid (no cascade). ``FM-field'': flow matching on the high-fidelity field rather than the residual.}
\label{tab:ablation}
\setlength{\tabcolsep}{4pt}
\begin{tabular}{lcccccccc}
\toprule
Method & Darcy & Burgers & SW & DR & Shear-T & Shear-P & AM  & NS \\
\midrule
MFFM-cascade      & \textbf{0.06198} & \textbf{0.02488} & 0.004756          & \textbf{0.2509} & \textbf{0.2253} & \textbf{0.2671} & \textbf{0.003162} & \textbf{0.04779} \\
w/o Gaussian blur & 0.07562          & 0.03591          & \textbf{0.004633} & 0.2701          & 0.2363          & 0.4358          & 0.003173          & 0.05224          \\
IID noise         & 0.1365           & 0.0396           & 0.02062           & 0.5081          & 0.4008          & 0.7120          & 0.003214          & 0.1151           \\
MFFM-single       & 0.08182          & 0.1091           & 0.04761           & 0.6871          & 0.5091          & 0.7057          & 0.003819          & 0.3136           \\
FM-field          & 0.5360           & 1.026            & 0.02062           & 0.5828          & 0.4848          & 0.5996          & 0.9578            & 0.3051           \\
\bottomrule
\end{tabular}
\end{table}

Each component contributes. Replacing the calibrated source with IID noise causes the largest and most consistent source-side degradation (e.g., $0.04779 \to 0.1151$ on Navier--Stokes, $0.2671 \to 0.7120$ on Shear-P), indicating that matching residual statistics matters more than matching average noise scale; Gaussian-blur correlation is a smaller but consistent additional gain on seven of eight benchmarks. Removing the cascade (MFFM-single) produces the largest drop on benchmarks with large LF-to-HF gaps (Burgers, SW, DR, NS). FM-field is substantially worse than residual MFFM on most benchmarks, confirming that flow matching is most effective here when applied to the unresolved bilinear residual rather than to the full high-fidelity field. A control variant that exposes the single-level model to multi-resolution training pairs (Table~\ref{tab:multires-control}, Appendix~\ref{app:multires-control}) does not close the gap to the full cascade, so the cascade gain is compositional rather than a side-effect of multi-resolution data exposure.

Ranking the four design choices by their effect on accuracy across the eight benchmarks: residual decomposition vs.\ field-space flow matching produces the largest swing, by an order of magnitude on Darcy, Burgers, and Active Matter; the cascade structure is the next largest contributor on the harder benchmarks; calibrated source statistics dominate on the easier benchmarks where the cascade gain saturates; Gaussian-blur correlation is the smallest of the four.

\subsection{Bilinear as the task-difficulty zero line}
\label{sec:exp-bilinear}

The bilinear baseline performs no learning: it upsamples $\uLF$ to the HF grid and reports the resulting NRMSE, which equals the relative $L^2$ norm of the residual $\norm{\delta} / \norm{\uHF}$. Reading the bilinear rows of Table~\ref{tab:main} as a task-difficulty axis organizes the benchmarks into a low-residual regime (Active Matter, Darcy, Shallow Water), an intermediate-residual regime (Burgers, Shear-T, Navier--Stokes), and a large-residual regime (Diffusion Reaction, Shear-P) where the LF solution is qualitatively wrong at the HF grid. Across this axis, MFFM remains competitive when bilinear is already strong, improves sharply when the LF-HF gap carries learnable local structure, and recovers useful predictions in the regime where several operator baselines degrade. Per-regime numerical breakdown appears in Appendix~\ref{app:bilinear-regimes}.

\subsection{Inference cost}
\label{sec:exp-nfe}

The deterministic MFFM-cascade used in Table~\ref{tab:main} performs one velocity-network evaluation per cascade level. This is the inference mode optimized by deterministic end-to-end fine-tuning: at each level the residual state is initialized at zero and one midpoint evaluation is applied. Thus, a cascade with $L$ refinement levels requires exactly $L$ velocity-network evaluations for point prediction.

Table~\ref{tab:nfe} (Appendix~\ref{app:nfe-table}) reports a representative NFE-scaling study in which trained checkpoints are evaluated with $\mathrm{NFE}/\text{level}\in\{1,2,5,10,50\}$. Increasing NFE does not improve NRMSE and only increases cost, because multi-step integration evaluates the velocity networks along off-training deterministic trajectories. We therefore use $\mathrm{NFE}/\text{level}=1$ for all main
deterministic results.

\section{Conclusion}
\label{sec:conclusion}



We presented Multi-Fidelity Flow Matching, a refinement framework for high-fidelity PDE solutions that conditions on a low-fidelity solve and uses a residual-calibrated source for flow-matching pretraining. Conditioning makes the LF-to-HF residual substantially easier to model than the full HF field, and the adapted source matches the scale and local structure of empirical PDE residuals. After independent level-wise pretraining, deterministic end-to-end cascade fine-tuning turns the model into a one-step-per-level refiner that reaches the finest grid in $L$ network evaluations per query.

Three caveats bound the scope. (i) MFFM requires paired LF and HF training data. (ii) The one-step operating point is optimized for deterministic refinement; stochastic dynamics or weaker LF-HF correlation may require richer source distributions or stochastic end-to-end training. (iii) Pretraining and inference use different conditioning chains; end-to-end fine-tuning addresses this empirically, and a formal stability guarantee is future work.

\newpage
\bibliography{references}

@inproceedings{lipman2023flow,
  title={Flow matching for generative modeling},
  author={Lipman, Yaron and Chen, Ricky T. Q. and Ben-Hamu, Heli and Nickel, Maximilian and Le, Matt},
  booktitle={International Conference on Learning Representations (ICLR)},
  year={2023}
}

@inproceedings{liu2023flow,
  title={Flow straight and fast: Learning to generate and transfer data with rectified flow},
  author={Liu, Xingchao and Gong, Chengyue and Liu, Qiang},
  booktitle={International Conference on Learning Representations (ICLR)},
  year={2023}
}

@article{tong2024improving,
  title={Improving and generalizing flow-based generative models with minibatch optimal transport},
  author={Tong, Alexander and Malkin, Nikolay and Huguet, Guillaume and Zhang, Yanlei and Rector-Brooks, Jarrid and Fatras, Kilian and Wolf, Guy and Bengio, Yoshua},
  journal={Transactions on Machine Learning Research (TMLR)},
  year={2024}
}

@article{albergo2023stochastic,
  title={Stochastic interpolants: A unifying framework for flows and diffusions},
  author={Albergo, Michael S. and Boffi, Nicholas M. and Vanden-Eijnden, Eric},
  journal={Journal of Machine Learning Research (JMLR)},
  volume={26},
  year={2025}
}

@article{lu2022multifidelity,
  title={Multifidelity deep neural operators for efficient learning of partial differential equations},
  author={Lu, Lu and Pestourie, Raphael and Johnson, Steven G. and Romano, Giuseppe},
  journal={Physical Review Research},
  year={2022}
}

@article{howard2023multifidelity,
  title={Multifidelity deep operator networks for data-driven and physics-informed problems},
  author={Howard, Amanda A. and Perego, Mauro and Karniadakis, George E. and Stinis, Panos},
  journal={Journal of Computational Physics},
  year={2023}
}

@article{kennedy2000predicting,
  title={Predicting the output from a complex computer code when fast approximations are available},
  author={Kennedy, Marc C. and O'Hagan, Anthony},
  journal={Biometrika},
  volume={87},
  number={1},
  pages={1--13},
  year={2000}
}

@article{perdikaris2017nonlinear,
  title={Nonlinear information fusion algorithms for data-efficient multi-fidelity modelling},
  author={Perdikaris, Paris and Raissi, Maziar and Damianou, Andreas and Lawrence, Neil D. and Karniadakis, George E.},
  journal={Proceedings of the Royal Society A},
  year={2017}
}

@article{peherstorfer2018survey,
  title={Survey of multifidelity methods in uncertainty propagation, inference, and optimization},
  author={Peherstorfer, Benjamin and Willcox, Karen and Gunzburger, Max},
  journal={SIAM Review},
  volume={60},
  number={3},
  pages={550--591},
  year={2018}
}

@inproceedings{li2021fourier,
  title={Fourier neural operator for parametric partial differential equations},
  author={Li, Zongyi and Kovachki, Nikola and Azizzadenesheli, Kamyar and Liu, Burigede and Bhattacharya, Kaushik and Stuart, Andrew and Anandkumar, Anima},
  booktitle={International Conference on Learning Representations (ICLR)},
  year={2021}
}

@article{lu2021learning,
  title={Learning nonlinear operators via DeepONet based on the universal approximation theorem of operators},
  author={Lu, Lu and Jin, Pengzhan and Pang, Guofei and Zhang, Zhongqiang and Karniadakis, George E.},
  journal={Nature Machine Intelligence},
  year={2021}
}

@inproceedings{bastek2025physics,
  title={Physics-informed diffusion models},
  author={Bastek, Jan-Hendrik and Sun, WaiChing and Kochmann, Dennis M.},
  booktitle={International Conference on Learning Representations (ICLR)},
  year={2025}
}

@inproceedings{baldan2025pbfm,
  title={Physics vs Distributions: {P}areto Optimal Flow Matching with Physics Constraints},
  author={Baldan, Giacomo and Liu, Qiang and Guardone, Alberto and Thuerey, Nils},
  booktitle={International Conference on Learning Representations (ICLR)},
  year={2026}
}

@inproceedings{hou2025cfo,
  title={{CFO}: Learning Continuous-Time {PDE} Dynamics via Flow-Matched Neural Operators},
  author={Hou, Xianglong and Huang, Xinquan and Perdikaris, Paris},
  booktitle={International Conference on Learning Representations (ICLR)},
  year={2026}
}

@inproceedings{huang2024diffusionpde,
  title={DiffusionPDE: Generative PDE-solving under partial observation},
  author={Huang, Jiahe and Yang, Guandao and Wang, Zichen and Park, Jeong Joon},
  booktitle={Advances in Neural Information Processing Systems (NeurIPS)},
  year={2024}
}

@inproceedings{lippe2024pde,
  title={PDE-Refiner: Achieving accurate long rollouts with neural PDE solvers},
  author={Lippe, Phillip and Veeling, Bas and Perdikaris, Paris and Turner, Richard E. and Brandstetter, Johannes},
  booktitle={Advances in Neural Information Processing Systems (NeurIPS)},
  year={2023}
}

@article{ho2022cascaded,
  title={Cascaded diffusion models for high fidelity image generation},
  author={Ho, Jonathan and Saharia, Chitwan and Chan, William and Fleet, David J. and Norouzi, Mohammad and Salimans, Tim},
  journal={Journal of Machine Learning Research (JMLR)},
  year={2022}
}

@inproceedings{saharia2022photorealistic,
  title={Photorealistic text-to-image diffusion models with deep language understanding},
  author={Saharia, Chitwan and Chan, William and Saxena, Saurabh and Li, Lala and Whang, Jay and Denton, Emily and Ghasemipour, Seyed Kamyar Seyed and Ayan, Burcu Karagol and Mahdavi, S. Sara and Lopes, Rapha Gontijo and others},
  booktitle={Advances in Neural Information Processing Systems (NeurIPS)},
  year={2022}
}

@inproceedings{debortoli2021bridge,
  title={Diffusion {S}chr\"odinger bridge with applications to score-based generative modeling},
  author={De Bortoli, Valentin and Thornton, James and Heng, Jeremy and Doucet, Arnaud},
  booktitle={Advances in Neural Information Processing Systems (NeurIPS)},
  year={2021}
}

@inproceedings{shi2024bridge,
  title={Diffusion {S}chr\"odinger bridge matching},
  author={Shi, Yuyang and De Bortoli, Valentin and Campbell, Andrew and Doucet, Arnaud},
  booktitle={Advances in Neural Information Processing Systems (NeurIPS)},
  year={2023}
}

@article{giles2015multilevel,
  title={Multilevel {M}onte {C}arlo methods},
  author={Giles, Michael B.},
  journal={Acta Numerica},
  volume={24},
  pages={259--328},
  year={2015}
}

@inproceedings{li2020multifidelity,
  title={Multi-fidelity {B}ayesian optimization via deep neural networks},
  author={Li, Shibo and Xing, Wei and Kirby, Robert and Zhe, Shandian},
  booktitle={Advances in Neural Information Processing Systems (NeurIPS)},
  year={2020}
}

@inproceedings{li2022deep,
  title={Deep multi-fidelity active learning of high-dimensional outputs},
  author={Li, Shibo and Wang, Zheng and Kirby, Robert and Zhe, Shandian},
  booktitle={International Conference on Artificial Intelligence and Statistics (AISTATS)},
  year={2022}
}

@inproceedings{li2022infinite,
  title={Infinite-fidelity coregionalization for physical simulation},
  author={Li, Shibo and Wang, Zheng and Kirby, Robert and Zhe, Shandian},
  booktitle={Advances in Neural Information Processing Systems (NeurIPS)},
  year={2022}
}

@inproceedings{li2024multiresolution,
  title={Multi-resolution active learning of {F}ourier neural operators},
  author={Li, Shibo and Yu, Xin and Xing, Wei and Kirby, Mike and Narayan, Akil and Zhe, Shandian},
  booktitle={International Conference on Artificial Intelligence and Statistics (AISTATS)},
  year={2024}
}

@misc{sun2025compol,
  title={{COMPOL}: A unified neural operator framework for scalable multi-physics simulations},
  author={Sun, Yifei and Wang, Tao and Qu, Junqi and Dong, Yushun and Tang, Hewei and Li, Shibo},
  year={2025},
  howpublished={arXiv preprint}
}

@inproceedings{pooladian2023multisample,
  title={Multisample flow matching: Straightening flows with minibatch couplings},
  author={Pooladian, Aram-Alexandre and Ben-Hamu, Heli and Domingo-Enrich, Carles and Amos, Brandon and Lipman, Yaron and Chen, Ricky T. Q.},
  booktitle={International Conference on Machine Learning (ICML)},
  year={2023}
}

@article{shu2023physics,
  title={A physics-informed diffusion model for high-fidelity flow field reconstruction},
  author={Shu, Dule and Li, Zijie and Farimani, Amir Barati},
  journal={Journal of Computational Physics},
  year={2023}
}

@inproceedings{greenfeld2019learning,
  title={Learning to optimize multigrid {PDE} solvers},
  author={Greenfeld, Daniel and Galun, Meirav and Basri, Ronen and Yavneh, Irad and Kimmel, Ron},
  booktitle={International Conference on Machine Learning (ICML)},
  year={2019}
}

@inproceedings{takamoto2022pdebench,
  title={{PDEBench}: An extensive benchmark for scientific machine learning},
  author={Takamoto, Makoto and Praditia, Timothy and Leiteritz, Raphael and MacKinlay, Daniel and Alesiani, Francesco and Pflüger, Dirk and Niepert, Mathias},
  booktitle={Advances in Neural Information Processing Systems (NeurIPS) Datasets and Benchmarks Track},
  year={2022}
}

@article{ohana2024well,
  title={The well: a large-scale collection of diverse physics simulations for machine learning},
  author={Ohana, Ruben and McCabe, Michael and Meyer, Lucas and Morel, Rudy and Agocs, Fruzsina J and Beneitez, Miguel and Berger, Marsha and Burkhart, Blakesley and Dalziel, Stuart B and Fielding, Drummond B and others},
  journal={Advances in Neural Information Processing Systems},
  volume={37},
  pages={44989--45037},
  year={2024}
}

@inproceedings{tran2023factorized,
  title={Factorized {F}ourier neural operators},
  author={Tran, Alasdair and Mathews, Alexander and Xie, Lexing and Ong, Cheng Soon},
  booktitle={International Conference on Learning Representations (ICLR)},
  year={2023}
}

@inproceedings{raonic2023convolutional,
  title={Convolutional neural operators for robust and accurate learning of {PDE}s},
  author={Raonic, Bogdan and Molinaro, Roberto and De Ryck, Tim and Rohner, Tobias and Bartolucci, Francesca and Alaifari, Rima and Mishra, Siddhartha and de B{\'e}zenac, Emmanuel},
  booktitle={Advances in Neural Information Processing Systems (NeurIPS)},
  year={2023}
}

@inproceedings{chen2018neural,
  title={Neural ordinary differential equations},
  author={Chen, Ricky T. Q. and Rubanova, Yulia and Bettencourt, Jesse and Duvenaud, David K.},
  booktitle={Advances in Neural Information Processing Systems (NeurIPS)},
  year={2018}
}

@inproceedings{ho2020denoising,
  title={Denoising diffusion probabilistic models},
  author={Ho, Jonathan and Jain, Ajay and Abbeel, Pieter},
  booktitle={Advances in Neural Information Processing Systems (NeurIPS)},
  year={2020}
}

@inproceedings{song2021scorebased,
  title={Score-based generative modeling through stochastic differential equations},
  author={Song, Yang and Sohl-Dickstein, Jascha and Kingma, Diederik P. and Kumar, Abhishek and Ermon, Stefano and Poole, Ben},
  booktitle={International Conference on Learning Representations (ICLR)},
  year={2021}
}

@inproceedings{karras2022elucidating,
  title={Elucidating the design space of diffusion-based generative models},
  author={Karras, Tero and Aittala, Miika and Aila, Timo and Laine, Samuli},
  booktitle={Advances in Neural Information Processing Systems (NeurIPS)},
  year={2022}
}

@article{kovachki2023neural,
  title={Neural operator: Learning maps between function spaces with applications to {PDE}s},
  author={Kovachki, Nikola and Li, Zongyi and Liu, Burigede and Azizzadenesheli, Kamyar and Bhattacharya, Kaushik and Stuart, Andrew and Anandkumar, Anima},
  journal={Journal of Machine Learning Research (JMLR)},
  volume={24},
  number={89},
  pages={1--97},
  year={2023}
}

@article{forrester2007multifidelity,
  title={Multi-fidelity optimization via surrogate modelling},
  author={Forrester, Alexander I. J. and S{\'o}bester, Andr{\'a}s and Keane, Andy J.},
  journal={Proceedings of the Royal Society A},
  volume={463},
  number={2088},
  pages={3251--3269},
  year={2007}
}

@inproceedings{brandstetter2022message,
  title={Message passing neural {PDE} solvers},
  author={Brandstetter, Johannes and Worrall, Daniel and Welling, Max},
  booktitle={International Conference on Learning Representations (ICLR)},
  year={2022}
}

@inproceedings{cachay2023dyffusion,
  title={{DY}ffusion: A dynamics-informed diffusion model for spatiotemporal forecasting},
  author={R{\"u}hling Cachay, Salva and Zhao, Bo and Joren, Hailey and Yu, Rose},
  booktitle={Advances in Neural Information Processing Systems (NeurIPS)},
  year={2023}
}

@inproceedings{holzschuh2023inverse,
  title={Solving inverse physics problems with score matching},
  author={Holzschuh, Benjamin and Vegetti, Simona and Thuerey, Nils},
  booktitle={Advances in Neural Information Processing Systems (NeurIPS)},
  year={2023}
}

\newpage
\appendix

\section{Conditioning and source-scale analysis}
\label{app:transport}

This appendix expands the mechanism discussed in Section~\ref{sec:method-onestep}. We separate
two roles that are coupled in MFFM: conditioning on the low-fidelity solution, which makes the
residual refinement problem structurally easier, and residual-calibrated source design, which
normalizes the flow-matching regression target. The main experiments use a deterministic
end-to-end fine-tuned cascade for point prediction; the analysis below explains why the residual
flow-matching pretraining provides a favorable initialization for that one-step operating point.

\subsection{Conditioning, not matched marginals, is the central mechanism}
\label{app:transport-mechanism}

The conditional flow-matching loss regresses the velocity network onto the per-sample conditional
velocity $\delta-\varepsilon$ evaluated at interpolated states
\[
    \delta_t = t\delta + (1-t)\varepsilon .
\]
The Bayes-optimal velocity represented by the trained network is the conditional regression
\[
    v^\star(z,t,\uLF)
    =
    \mathbb{E}[\delta-\varepsilon \mid \delta_t=z,\; t,\; \uLF].
\]

The key structural simplification comes from conditioning on $\uLF$. In the deterministic
multi-fidelity setting considered here, the high-fidelity solution is a deterministic function of the
underlying PDE instance. If the low-fidelity solution identifies that instance sufficiently well, then
the conditional law of the residual collapses around a deterministic refinement target
\[
    g(\uLF) := \uHF - \uLF .
\]
In the idealized injective case, this conditional law is a Dirac measure at $g(\uLF)$. Given
$\uLF$ and $t<1$, the source can be recovered from a point on the interpolation path by
\[
    \varepsilon = \frac{\delta_t - t\,g(\uLF)}{1-t}.
\]
Substituting this identity into the target $\delta-\varepsilon$ gives the conditional Bayes regressor
\[
    v^\star(z,t,\uLF)
    =
    \frac{g(\uLF)-z}{1-t}.
\]
Along the stochastic interpolation path initialized at $\delta_0=\varepsilon$, this velocity reduces
to the constant direction $g(\uLF)-\varepsilon$. This idealized calculation explains why a strongly
conditioned residual flow is much easier than unconditional field generation.

The unconditional marginal velocity does not have this property. For example, if $\delta$ and
$\varepsilon$ are independent zero-mean Gaussians with equal covariance, then
\begin{equation}
v^\star_{\mathrm{marginal}}(z,t)
=
\mathbb{E}[\delta-\varepsilon \mid \delta_t=z,t]
=
\frac{2t-1}{2t^2-2t+1}\,z,
\end{equation}
which is explicitly time-dependent. The conditional and marginal calculations are not in conflict:
the former exploits the information in $\uLF$, whereas the latter does not.

In the main experiments, the final point predictor is not the stochastic path above. After level-wise
flow-matching pretraining, we fine-tune the composed cascade using a deterministic rollout with
zero residual initialization and one midpoint velocity evaluation per level. Thus, the calculation
above should be read as a conditioning argument for why residual flow matching is a good
pretraining objective, not as a claim that the final deterministic e2e predictor is a classical
many-step ODE solver.

\subsection{Residual-calibrated sources as normalization}
\label{app:transport-marginal}

The source distribution in MFFM is a design choice. The diagonal residual source
\[
    \varepsilon_{\mathrm{diag}} \sim
    \mathcal{N}\!\left(0,\mathrm{diag}(\sigma_\delta^2)\right),
    \qquad
    \sigma_{\delta,j}^2=\widehat{\mathrm{Var}}_{\mathcal{D}}[\delta_j],
\]
matches the empirical per-coordinate residual scale. For analytic clarity, consider this diagonal
case. If $\varepsilon_{\mathrm{diag}}$ is independent of $\delta$ and has the same per-coordinate
variance as the empirical residual, then the regression target satisfies
\[
    \mathbb{E}\|\delta-\varepsilon_{\mathrm{diag}}\|_2^2
    =
    2\,\mathrm{Tr}\!\left(\mathrm{diag}(\sigma_\delta^2)\right).
\]
This avoids the $\Theta(n)$ target scale induced by an isotropic unit-variance source on
unit-normalized fields and keeps the flow-matching regression problem numerically well scaled.

The main implementation further applies local Gaussian-blur correlation. Concretely, we sample
iid noise, apply a depthwise Gaussian blur, normalize each sample to unit empirical standard
deviation, and then scale by $\sigma_\delta$. This preserves the residual scale used in the diagonal
analysis while adding local spatial correlation, which better matches the structure of PDE residual
fields. The ablations in Section~\ref{sec:exp-ablation} show that removing Gaussian blur or replacing
the residual-calibrated source with IID noise degrades accuracy.

\subsection{Per-coordinate variance reduction}
\label{app:transport-variance}

Let $\rho_j=\mathrm{Corr}(\uLF[j],\uHF[j])$ denote the empirical correlation between the two
fidelities at coordinate $x_j$. The residual variance is
\begin{equation}
\sigma_{\delta,j}^2
=
\mathrm{Var}[\uHF[j]-\uLF[j]]
=
\sigma_{\mathrm{HF},j}^2
+
\sigma_{\mathrm{LF},j}^2
-
2\rho_j\sigma_{\mathrm{HF},j}\sigma_{\mathrm{LF},j}.
\end{equation}
When the two fidelities have similar per-coordinate variance
($\sigma_{\mathrm{LF},j}\approx\sigma_{\mathrm{HF},j}$), this simplifies to
\begin{equation}
\sigma_{\delta,j}^2
\approx
2(1-\rho_j)\sigma_{\mathrm{HF},j}^2 .
\label{eq:appA-residual-var}
\end{equation}
Table~\ref{tab:appA-rho} reports the corresponding variance ratio for representative correlations.

\begin{table}[t]
\centering
\caption{
Per-coordinate residual variance ratio $\sigma_\delta^2/\sigma_{\mathrm{HF}}^2$ from
Eq.~\eqref{eq:appA-residual-var}, as a function of empirical correlation $\rho$ between fidelities.
}
\label{tab:appA-rho}
\begin{tabular}{lccccc}
\toprule
$\rho$ & 0.50 & 0.70 & 0.90 & 0.95 & 0.99 \\
\midrule
$\sigma_{\delta}^2 / \sigma_{\mathrm{HF}}^2$ & 1.00 & 0.60 & 0.20 & 0.10 & 0.02 \\
\bottomrule
\end{tabular}
\end{table}

This calculation explains why adjacent-fidelity residual refinement is easier than full-field
generation when fidelities are strongly correlated. It also motivates the cascade: adjacent levels
typically have higher correlation than the coarsest-to-finest pair, reducing the scale of the residual
that each level must learn.

\subsection{Relation to deterministic one-step inference}
\label{app:transport-onestep}

The analysis above does not imply that the final e2e fine-tuned model should be evaluated with
many ODE steps. In fact, the main deterministic predictor is explicitly trained as a one-step-per-level
cascade. At level $\ell$, the rollout used during e2e fine-tuning is
\[
    \hat u_{\ell+1}
    =
    I_{\ell\to\ell+1}(\hat u_\ell)
    +
    v_{\theta_\ell}\!\left(
        0,\frac{1}{2},I_{\ell\to\ell+1}(\hat u_\ell)
    \right).
\]
Therefore, increasing the number of Euler evaluations at test time changes the operating
distribution of the velocity networks. The deterministic NFE-scaling study in
Appendix~\ref{app:nfe-table} confirms that larger NFE does not improve the fine-tuned predictor.
This is why MFFM-cascade is used as a one-step-per-level deterministic refiner in all main results.
\section{Implementation details}
\label{app:impl}

This appendix collects the architectural, data-construction, and training-protocol details elided from Section~\ref{sec:method-arch}.

\subsection{Per-benchmark configuration}
\label{app:impl-bench}

Table~\ref{tab:appB-train-config} lists the cascade resolutions, train/validation/test split sizes, and training-stage epoch budgets used for each benchmark. The same shared training protocol described in Section~\ref{sec:exp-setup} (AdamW, cosine schedule, gradient clipping, EMA) is used throughout; the only per-benchmark variation is the resolution hierarchy, the split sizes, and the epoch budgets reported here.

\begin{table}[t]
\centering
\caption{
Per-benchmark data and training configuration. ``Split'' reports train/validation/test sample counts. ``Pretrain epochs'' denotes independent level-wise flow-matching training. ``E2E FT epochs'' denotes deterministic end-to-end cascade fine-tuning after level-wise pretraining. Cascade resolutions list the grids visited by the level-wise refinement.
}
\label{tab:appB-train-config}
\setlength{\tabcolsep}{4pt}
\resizebox{\textwidth}{!}{%
\begin{tabular}{lccccc}
\toprule
Benchmark & Task type & Cascade resolutions & Split & Pretrain epochs & E2E FT epochs \\
\midrule
Darcy & spatial SR & $\{16,32,64,128\}^2$ & 512/128/128 & 500 & 100 \\
Burgers & space--time SR & $\{32,64,128,256\}^2$ & 512/128/128 & 100 & 100 \\
Shallow Water & block forecasting & $\{16,32,64,128\}^2$ & 800/100/100 & 100 & 100 \\
Diffusion Reaction & block forecasting & $\{16,32,64,128\}^2$ & 800/100/100 & 300 & 300 \\
Shear Flow (tracer) & block forecasting & $\{16,32,64,128\}^2$ & 896/112/112 & 300 & 300 \\
Shear Flow (pressure) & block forecasting & $\{16,32,64,128\}^2$ & 896/112/112 & 300 & 300 \\
Active Matter & block forecasting & $\{16,32,64,128\}^2$ & 175/24/26 & 100 & 100 \\
Navier--Stokes & block forecasting & $\{8,16,32,64\}^2$ & 2048/256/256 & 300 & 100 \\
\bottomrule
\end{tabular}
}
\end{table}

\subsection{Data construction}
\label{app:impl-data}

We use two data-construction protocols: super-resolution of paired solution fields and spatiotemporal block forecasting. In all cases, lower-resolution fields are bilinearly prolonged to the target resolution before forming residuals.

\paragraph{Darcy.}
The Darcy benchmark is generated in-house as a multi-resolution elliptic PDE dataset. We sample $768$ coefficient fields from a Gaussian random field prior and construct lognormal permeability fields. For each sample, the same underlying high-resolution coefficient field is downsampled to the
cascade resolutions $\{16,32,64,128\}^2$, and the Darcy equation is solved separately on each grid. Thus, $u_{16},u_{32},u_{64}$, and $u_{128}$ are paired multi-resolution solver outputs for the same underlying PDE instance, rather than resized copies of the high-resolution solution. The resulting paired fields are split into $512/128/128$ train/validation/test samples. The cascade learns spatial residual refinements over the hierarchy $\{16,32,64,128\}^2$.

\paragraph{Burgers.}
The Burgers benchmark is generated in-house from the one-dimensional viscous Burgers equation. Although the underlying PDE is one-dimensional in space, we represent each solution trajectory as a two-dimensional space--time field $u(t,x)$. For each of $768$ samples, we draw one continuous
Gaussian-random-field initial condition and solve the same underlying sample at resolutions $\{32,64,128,256\}$. At resolution $s$, we use $s$ spatial points and save $s$ future time snapshots, giving an output tensor of shape $s \times s$. Thus, both the spatial and temporal axes are included in the resolution hierarchy, and the task is space--time super-resolution rather than one-step temporal forecasting. The data are split into $512/128/128$ train/validation/test samples.

\paragraph{Spatiotemporal forecasting benchmarks.}
For Shallow Water, Diffusion Reaction, Shear Flow, Active Matter, and Navier--Stokes, we use block-to-block forecasting. Each sample contains $50$ frames. The first $25$ frames are used as the coarse-grid input block, and the last $25$ frames are used as the high-fidelity target block. The
temporal axis is stacked into the channel dimension, so a block with $25$ frames and $C_{\mathrm{phys}}$ physical variables is represented as a tensor with $25C_{\mathrm{phys}}$ channels. Shallow Water and Diffusion Reaction are taken from PDEBench; Shear Flow and Active Matter are taken from The Well; Navier--Stokes is taken from the FNO dataset \texttt{PDENavierStokes\_V1e-3\_N5000\_T50}. When the native spatial grid is not square, we resize the fields to square grids before constructing the cascade resolutions. The level-wise residual at cascade level $\ell$ is written in a form that covers both super-resolution and past-to-future block forecasting:
\[
    \delta_\ell
    =
    u_{\ell+1}^{\mathrm{tar}}
    -
    \Icoarsefine{\ell}{\ell+1}(c_\ell),
\]
where $u_{\ell+1}^{\mathrm{tar}}$ denotes the target field or target frame block at resolution level $\ell+1$, and $c_\ell$ denotes the conditioning field at level $\ell$. For $\ell=0$, $c_0$ is the observed coarse input field or coarse past-frame block. For later cascade levels, $c_\ell$ is the lower-resolution target field during level-wise pretraining and the prediction propagated from the previous cascade level during inference.

For Darcy and Burgers, the conditioning field is simply the lower-resolution solution field, $c_\ell = u_\ell$, so the residual reduces to
\[
    \delta_\ell
    =
    u_{\ell+1}
    -
    \Icoarsefine{\ell}{\ell+1}(u_\ell).
\]
For the spatiotemporal forecasting benchmarks, $u_{\ell+1}^{\mathrm{tar}}$ is the stacked future-frame target block, while $c_\ell$ is the corresponding conditioning block used at that cascade level.

\subsection{Architecture rationale}
\label{app:impl-rationale}

The velocity network is a U-Net-style residual convolutional network mapping the channel concatenation $(\delta_t, \uLF) \in \R^{B \times 2C \times H \times W}$ to a velocity field in $\R^{B \times C \times H \times W}$ on the same lattice. Here $C$ denotes the number of solution channels, equal to $T \cdot C_{\mathrm{phys}}$ in the spatiotemporal forecasting setting where time frames are stacked as channels. The network uses local convolutions, residual blocks, skip additions, and time-dependent FiLM conditioning. We use this lightweight image-to-image backbone because the architectural choice is not the main contribution; the method only requires a velocity network that takes $(\delta_t,\uLF,t)$ and outputs a residual velocity on the target grid.

The architectural extension specific to MFFM is the channel concatenation of $\uLF$ with $\delta_t$ at the network input, which exposes the cheap solution to all convolutional layers. We use channel concatenation rather than a global conditioning bottleneck because $\uLF$ is a structured spatial field rather than a low-dimensional descriptor. The cascade prolongation $\Icoarsefine{\ell}{\ell+1}$ is bilinear interpolation; learned prolongations would couple cascade levels and break the simple level-wise construction.

\subsection{Velocity network}
\label{app:impl-net}

Each per-level velocity network $\vtheta$ has the following components.

\paragraph{Input and output.}
The input is the channel concatenation $(\delta_t, \uLF) \in \R^{B \times 2C \times H \times W}$, where $\delta_t$ is the interpolated residual state and $\uLF$ denotes the prolongated conditioning field at the current level. The output is a velocity field $\vtheta(\delta_t,t,\uLF) \in \R^{B \times C \times H \times W}$.

\paragraph{Input lift.}
A $3{\times}3$ convolution maps the $2C$ input channels to a hidden width $h$.

\paragraph{Time conditioning.}
The scalar time $t \in [0,1]$ is embedded with a sinusoidal embedding and passed through a two-layer MLP to produce a hidden time-conditioning vector. Each residual block receives this conditioning vector through a FiLM module.

\paragraph{Residual blocks and skip additions.}
The network applies a stack of residual convolutional blocks followed by a second stack with skip additions from the first stack. Each block uses group normalization, SiLU activations, two $3{\times}3$ convolutions, and FiLM modulation from the time embedding. All convolutions operate at the current target-grid resolution; the velocity network itself does not perform spatial downsampling or upsampling.

\paragraph{Output head.}
A final group normalization, SiLU activation, and zero-initialized $1{\times}1$ convolution map the hidden features back to $C$ channels. Zero initialization makes the initial velocity approximately zero, giving a stable identity-like initialization for the ODE update.

\paragraph{Group normalization.}
We use group normalization with up to eight groups, choosing a divisor of the hidden width when necessary.

\paragraph{Exponential moving average.}
During independent level-wise flow-matching pretraining, model weights are tracked by an EMA copy with decay $0.999$, and the EMA weights are applied before the cascade fine-tuning stage. The deterministic end-to-end fine-tuning stage does not use EMA; the final checkpoint is selected by validation NRMSE.

\subsection{Training protocol}
\label{app:impl-train}

\paragraph{Optimizer.} Independent level-wise flow-matching pretraining uses AdamW with learning rate $10^{-5}$, $\beta_1 = 0.9$, $\beta_2 = 0.999$, weight decay $10^{-6}$, and gradient clipping at L2 norm $1.0$. EMA weights with decay $0.999$ are used during this pretraining stage. After level-wise pretraining, MFFM-cascade is fine-tuned end-to-end using the deterministic cascade rollout with one fifth of the base learning rate, the same weight decay, and the same gradient clipping. The end-to-end fine-tuned checkpoint is selected by validation NRMSE.

\paragraph{Schedule.} Cosine learning-rate decay over the full training horizon, no warmup at small scale.

\paragraph{Batch and epoch counts.} Batch size is 32. Per-benchmark settings appear in Table~\ref{tab:appB-train-config}.

\paragraph{Source-statistics precomputation.}
Before training each level, the empirical per-coordinate residual variance $\sigma_j^2 = \widehat{\Var}_{\Dcal}[\delta_j]$ is computed on the corresponding training residuals and stored. For cascade level $\ell$, the residual is
\[
    \delta_\ell
    =
    u_{\ell+1}^{\mathrm{tar}}
    -
    \Icoarsefine{\ell}{\ell+1}(c_\ell).
\]
For pure super-resolution tasks, this reduces to $\delta_\ell=u_{\ell+1}-\Icoarsefine{\ell}{\ell+1}(u_\ell)$. The source noise is generated by
sampling iid Gaussian noise, applying a depthwise Gaussian blur kernel, normalizing each sample to unit empirical standard deviation, and scaling by the stored residual standard deviation $\sigma_\delta$. The variance is floored at $10^{-8}$ to avoid numerical issues at coordinates with
near-zero residual variance.

\paragraph{Per-level width and depth.}
The hidden width and number of residual blocks are selected per cascade level. Earlier levels usually handle larger LF-to-HF corrections, while later levels mainly perform smaller high-resolution refinements, so later levels are assigned no larger capacity than earlier ones. In the implementation used for the main experiments, the hidden width is lower-bounded by $32$ and rounded to a multiple of eight, and the number of residual blocks is lower-bounded by two. Exact per-level values are recorded in the released training scripts.

\subsection{Algorithm pseudocode}
\label{app:impl-algorithms}

\begin{algorithm}[t]
\caption{MFFM training (single cascade level)}
\label{alg:train}
\begin{algorithmic}[1]
\Require Level-wise paired data
$\Dcal_\ell = \{(\tilde u_\ell^{(i)}, u_{\ell+1}^{\mathrm{tar},(i)})\}_{i=1}^N$,
network $\vtheta$, total steps $T$
\State $\delta^{(i)} \gets u_{\ell+1}^{\mathrm{tar},(i)} - \tilde u_\ell^{(i)}$ for all training pairs
\State $\sigma_\delta^2 \gets \widehat{\Var}_{\Dcal_\ell}[\delta]$ \Comment{per-coordinate residual variance}
\For{step $= 1$ to $T$}
  \State Sample minibatch $(\tilde u_\ell, u_{\ell+1}^{\mathrm{tar}}) \sim \Dcal_\ell$
  \State $\delta \gets u_{\ell+1}^{\mathrm{tar}} - \tilde u_\ell$
  \State Sample $z \sim \N(0,I)$
  \State $\bar z \gets K_\tau * z$ \Comment{depthwise Gaussian blur}
  \State $\tilde z \gets \bar z / (\widehat{\mathrm{Std}}(\bar z) + \epsilon_{\mathrm{num}})$
  \State $\varepsilon \gets \sigma_\delta \odot \tilde z$
  \State Sample $t \sim \mathcal{U}[0,1]$
  \State $\delta_t \gets t\,\delta + (1-t)\,\varepsilon$
  \State $\mathcal{J} \gets \norm{\vtheta(\delta_t, t, \tilde u_\ell) - (\delta - \varepsilon)}^2$
  \State Update $\theta$ by AdamW on $\mathcal{J}$; update EMA weights
\EndFor
\State Apply EMA weights to $\vtheta$
\State \Return $\vtheta$
\end{algorithmic}
\end{algorithm}

Algorithm~\ref{alg:cascade-det} gives the deterministic rollout used both for point prediction and
for end-to-end cascade fine-tuning. During fine-tuning, the returned $\hat u_L$ is compared with the
finest-resolution target, and all level networks are optimized jointly.

\begin{algorithm}[t]
\caption{Deterministic cascade point prediction: produces $\hat u_L$ at the finest grid using one velocity-network evaluation per cascade level.}
\label{alg:cascade-det}
\begin{algorithmic}[1]
\Require Coarse-grid input $u_0$, trained per-level networks $\{v_{\theta_\ell}\}_{\ell=0}^{L-1}$
\State $\hat u_0 \gets u_0$
\For{$\ell = 0$ to $L-1$}
  \State $\tilde u_\ell \gets \Icoarsefine{\ell}{\ell+1}(\hat u_\ell)$
  \State $\delta_\ell \gets 0$
  \State $v_\ell \gets v_{\theta_\ell}(\delta_\ell,\, 1/2,\, \tilde u_\ell)$
  \State $\hat u_{\ell+1} \gets \tilde u_\ell + v_\ell$
\EndFor
\State \Return $\hat u_L$
\end{algorithmic}
\end{algorithm}
\section{Additional empirical analysis}
\label{app:extra-ablations}

\subsection{Bilinear-axis regime breakdown}
\label{app:bilinear-regimes}

This appendix expands the per-regime discussion summarized in Section~\ref{sec:exp-bilinear}.
The bilinear baseline measures the relative size of the LF--HF residual before learning,
\[
    \mathrm{NRMSE}_{\mathrm{bilinear}}
    =
    \frac{\|u_{\mathrm{HF}} - I(u_{\mathrm{LF}})\|_2}
    {\|u_{\mathrm{HF}}\|_2}.
\]
It therefore provides a simple axis for interpreting where refinement methods have room to improve.

\paragraph{Low-residual regime.}
Active Matter ($0.0036$), Darcy ($0.0862$), and Shallow Water ($0.1007$) have relatively small bilinear residuals. In this regime, the LF input already captures much of the HF structure, although Shallow Water still contains learnable high-frequency error: MFFM reduces its NRMSE from
$0.1007$ to $0.004756$. On Active Matter and Darcy, the absolute headroom is smaller, and the difference between strong refinement methods is correspondingly modest.

\paragraph{Intermediate-residual regime.}
Burgers ($0.1541$), Shear Flow tracer ($0.6260$), and Navier--Stokes ($0.6172$) require nontrivial local correction beyond interpolation. MFFM improves these to $0.02488$, $0.2253$, and $0.04779$, respectively, indicating that the cascade can recover unresolved local structure from the LF input.

\paragraph{Large-residual regime.}
Diffusion Reaction ($3.170$) and Shear Flow pressure ($0.9011$) have the largest bilinear residuals. These are the settings where direct residual learning and operator baselines are most unstable. MFFM reduces the errors to $0.2509$ and $0.2671$, respectively, supporting the value of splitting the LF--HF correction into adjacent-fidelity residual refinements.

\subsection{Deterministic NFE scaling}
\label{app:nfe-table}

The main MFFM-cascade results use the deterministic one-step rollout optimized during end-to-end fine-tuning. This appendix tests whether increasing the number of Euler evaluations per cascade level improves the fine-tuned deterministic predictor. For each benchmark, we train a representative single-seed checkpoint with $\mathrm{NFE}/\text{level}=1$ during deterministic end-to-end fine-tuning, then keep all network weights fixed and vary only $\mathrm{NFE}/\text{level}\in\{1,2,5,10,50\}$ at evaluation time. The residual state is initialized at zero at each level, matching the deterministic point-prediction mode used in the main experiments. A larger NFE therefore does not correspond to additional training or a different model; it only changes the numerical rollout used at inference.

\begin{table}[h]
\centering
\small
\caption{
Deterministic NFE-scaling study for MFFM-cascade. Each row uses one representative single-seed
checkpoint trained with $\mathrm{NFE}/\text{level}=1$ during deterministic end-to-end fine-tuning;
only the number of Euler evaluations per cascade level is changed at evaluation time. The one-step
setting is therefore the optimized operating point. Total NFE is
$\mathrm{NFE}/\text{level} \times L$, with $L=3$ refinement levels for the listed benchmarks.
Table~\ref{tab:main} reports multi-seed averages for the main comparison.
}
\label{tab:nfe}
\setlength{\tabcolsep}{6pt}
\begin{tabular}{lccccc}
\toprule
Benchmark / NFE per level & 1 & 2 & 5 & 10 & 50 \\
\midrule
Total NFE & 3 & 6 & 15 & 30 & 150 \\
Burgers & \textbf{0.02556} & 0.03794 & 0.04660 & 0.04900 & 0.05081 \\
Shallow Water & \textbf{0.003378} & 0.008787 & 0.01195 & 0.01297 & 0.01376 \\
Active Matter & \textbf{0.003082} & \textbf{0.003082} & 0.003084 & 0.003085 & 0.003085 \\
Navier--Stokes & \textbf{0.05996} & 0.1054 & 0.1363 & 0.1458 & 0.1531 \\
\bottomrule
\end{tabular}
\end{table}

Across the tested benchmarks, increasing NFE does not improve the fine-tuned deterministic predictor. The best or tied-best NRMSE is obtained at $\mathrm{NFE}/\text{level}=1$, and larger values either leave the error nearly unchanged or degrade it. This behavior is consistent with the training procedure: end-to-end fine-tuning optimizes the composed one-step rollout directly, while larger NFE evaluates the velocity networks along deterministic trajectories that were not optimized by the fine-tuning objective.

The result should therefore be interpreted as an operating-point study rather than as a classical ODE solver convergence test. MFFM-cascade is designed to be used as a one-step-per-level deterministic refiner after end-to-end fine-tuning. Additional Euler evaluations increase the number of velocity-network calls linearly without improving the point predictor.




\subsection{Multi-resolution training pairs are not the source of the cascade gain}
\label{app:multires-control}

Section~\ref{sec:exp-ablation} compares the full cascade against a single-level model trained from coarsest to finest. A natural concern is that the cascade improvement reflects exposure to multi-resolution training pairs rather than a compositional benefit of adjacent-fidelity refinement. We control for this with \textsc{MFFM-single (multires train)}: a single residual flow trained on the same multi-resolution pairs available to the cascade, but evaluated as a direct coarsest-to-finest refiner.

\begin{table}[h]
\centering
\small
\caption{Single-level model with multi-resolution training data does not match the full cascade. Numbers are global NRMSE; \textsc{MFFM-cascade} and \textsc{MFFM-single} repeat values from Table~\ref{tab:ablation} for reference.}
\label{tab:multires-control}
\setlength{\tabcolsep}{6pt}
\resizebox{\textwidth}{!}{%
\begin{tabular}{lcccccccc}
\toprule
Method & Darcy & Burgers & SW & DR & Shear-T & Shear-P & AM & NS \\
\midrule
MFFM-cascade                  & 0.06198 & 0.02488 & 0.004756 & 0.2509 & 0.2253 & 0.2671 & 0.003162 & 0.04779 \\
MFFM-single                   & 0.08182 & 0.1091  & 0.04761  & 0.6871 & 0.5091 & 0.7057 & 0.003819 & 0.3136  \\
MFFM-single (multires train)  & 0.08136 & 0.1227  & 0.01702  & 0.5591 & 0.5892 & 0.5023 & 0.04354  & 0.2624  \\
\bottomrule
\end{tabular}
}
\end{table}

The multires-trained single-level model improves over the ordinary single-level model on Shallow Water and Diffusion Reaction but remains far worse than the cascade; on Shear-T and Active Matter it is even worse than the ordinary single-level variant. The cascade's compositional adjacent-resolution refinement is therefore the load-bearing structural choice, not multi-resolution data exposure alone.

\subsection{Stochastic end-to-end cascade variant}
\label{app:stochastic-e2e}

The main MFFM-cascade results in Table~\ref{tab:main} use deterministic end-to-end fine-tuning and deterministic point prediction. We additionally test a stochastic end-to-end variant to assess whether the source-sampling interface can be optimized directly. This experiment is not used for the main accuracy comparison; it is included as an appendix diagnostic.

Starting from the same level-wise residual flow-matching pretraining, we fine-tune the composed cascade with stochastic rollouts. At each cascade level, the residual state is initialized by a source sample
\[
    \delta_{\ell,0} = \varepsilon_\ell, 
    \qquad
    \varepsilon_\ell \sim q_{\delta_\ell},
\]
and the learned velocity field is integrated with one Euler evaluation per level. For a minibatch, we draw $K_{\mathrm{train}}$ independent stochastic rollouts and optimize the Monte Carlo estimate of the expected relative error:
\[
    \mathcal{L}_{\mathrm{stoch\text{-}e2e}}
    =
    \frac{1}{K_{\mathrm{train}}}
    \sum_{k=1}^{K_{\mathrm{train}}}
    \frac{
        \|\hat u_L^{(k)} - u_L\|_2
    }{
        \|u_L\|_2 + \epsilon
    }.
\]
In the implementation used here, $K_{\mathrm{train}}=4$ and $K_{\mathrm{val}}=8$. At test time, we draw $K_{\mathrm{test}}=10$ samples and report the NRMSE of the ensemble mean together with the average predictive spread:
\[
    \bar u_L = \frac{1}{K_{\mathrm{test}}}\sum_{k=1}^{K_{\mathrm{test}}}\hat u_L^{(k)},
    \qquad
    s = \mathrm{mean}_x \, \mathrm{Std}_k[\hat u_L^{(k)}(x)].
\]

\begin{table}[t]
\centering
\small
\caption{
Stochastic end-to-end cascade variant. The model is fine-tuned using stochastic cascade rollouts and evaluated by the ensemble mean over $K_{\mathrm{test}}=10$ samples. ``Spread'' is the mean pointwise standard deviation across stochastic samples. This experiment is a diagnostic of the stochastic sampling interface and is not used for the main deterministic comparison.
}
\label{tab:stochastic-e2e}
\setlength{\tabcolsep}{5pt}
\resizebox{\textwidth}{!}{%
\begin{tabular}{lcccc}
\toprule
Benchmark & NRMSE (NFE/level=1) & Spread (NFE/level=1) & NRMSE (NFE/level=2) & Spread (NFE/level=2) \\
\midrule
Darcy             & 0.07701 & $2.232{\times}10^{-4}$ & \textbf{0.07693} & $2.233{\times}10^{-4}$ \\
Burgers           & \textbf{0.03274} & $2.209{\times}10^{-3}$ & 0.04089 & $2.955{\times}10^{-3}$ \\
Shallow Water     & \textbf{0.01207} & $1.012{\times}10^{-2}$ & 0.02815 & $9.940{\times}10^{-3}$ \\
Diffusion Reaction & \textbf{0.3325} & $1.269{\times}10^{-2}$ & 0.3776 & $1.182{\times}10^{-2}$ \\
Shear-P           & \textbf{0.3052} & $4.888{\times}10^{-3}$ & 0.5339 & $4.271{\times}10^{-3}$ \\
Shear-T           & \textbf{0.2524} & $2.897{\times}10^{-2}$ & 0.4175 & $2.300{\times}10^{-2}$ \\
Active Matter     & \textbf{0.003226} & $1.768{\times}10^{-3}$ & 0.003228 & $1.768{\times}10^{-3}$ \\
Navier--Stokes    & \textbf{0.05711} & $4.057{\times}10^{-2}$ & 0.1227 & $3.303{\times}10^{-2}$ \\
\bottomrule
\end{tabular}
}
\end{table}

The stochastic variant confirms that the learned residual flow can support a sampling-based predictor, but it is not the operating point used for the main results. In most benchmarks, the stochastic ensemble mean is less accurate than the deterministic e2e predictor in Table~\ref{tab:main}, especially when moving from one to two Euler evaluations per level. This mirrors the deterministic NFE-scaling behavior in Appendix~\ref{app:nfe-table}: the networks are most effective at the one-step operating point for which the cascade is fine-tuned. We therefore use deterministic e2e prediction for the headline NRMSE comparison and treat stochastic rollouts as an optional diagnostic of predictive spread.

\subsection{Visualization}
\label{app:ns-vis}

\begin{figure}[t]
\centering
\includegraphics[width=\textwidth]{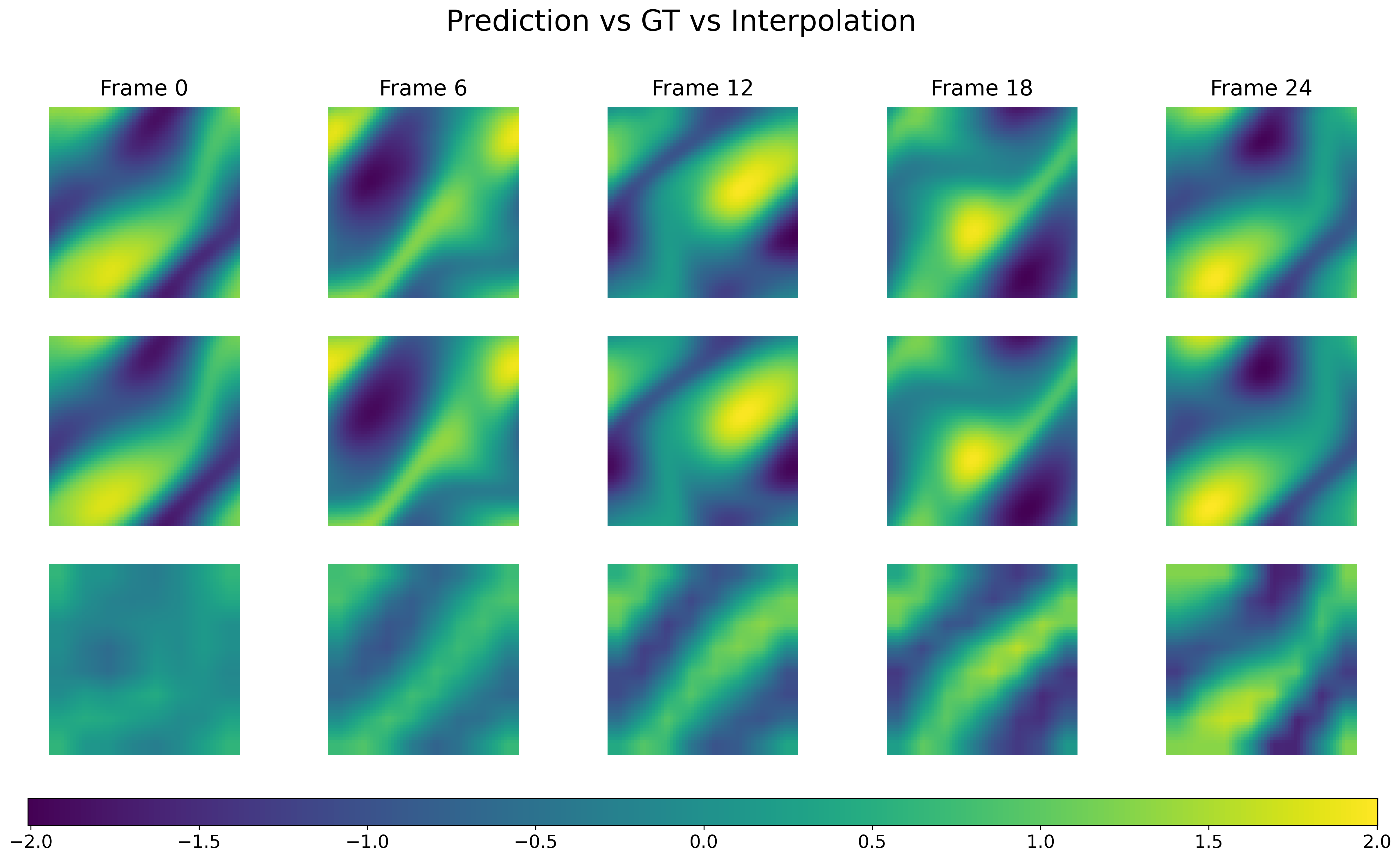}
\caption{
Qualitative Navier--Stokes visualization on a held-out test trajectory. Columns show five selected future frames. Rows show MFFM prediction, high-fidelity ground truth, and bilinear interpolation, respectively. MFFM preserves the large-scale structure while correcting local errors left by pure
interpolation.
}
\label{fig:ns-vis-app}
\end{figure}

Figure~\ref{fig:ns-vis-app} provides a qualitative example on the Navier--Stokes benchmark. We visualize five predicted future frames from one held-out test trajectory. The rows show the MFFM cascade prediction, the high-fidelity ground truth, and the bilinear interpolation baseline, respectively. The columns correspond to representative future-frame indices. This figure is intended as a qualitative diagnostic rather than a separate quantitative claim; the full numerical comparison is reported in Table~\ref{tab:main}.

The example illustrates the role of learned residual refinement. Bilinear interpolation captures the large-scale flow pattern but produces overly smooth structures and misses sharper local variations. The MFFM cascade prediction more closely matches the high-fidelity target across the selected
frames, consistent with the lower Navier--Stokes NRMSE reported in Table~\ref{tab:main}.

\end{document}